\documentclass[runningheads]{llncs}

\usepackage{eccv}

\usepackage{eccvabbrv}

\usepackage{graphicx}
\usepackage{booktabs}

\usepackage[accsupp]{axessibility}  
\usepackage{tabularx}
\usepackage{multirow}
\usepackage{wrapfig}
\usepackage{tabu}
\usepackage{arydshln}
\usepackage{amssymb}
\usepackage{float}
\usepackage[list=true]{subcaption}

\newcommand{\R}{\ensuremath{\mathbb{R}}\xspace}

\newcommand{\vx}{\ensuremath{\vec{x}}\xspace}
\newcommand{\vw}{\ensuremath{\vec{w}}\xspace}

\DeclareMathOperator{\Tr}{Tr}
\newcommand{\DinoFD}{{FD}$_\text{DINO-V2}$\xspace}

\usepackage{hyperref}

\usepackage{orcidlink}

\begin{document}
\title{Enhancing Plausibility Evaluation for Generated Designs with Denoising Autoencoder} 

\titlerunning{Fr\'echet Denoised Distance}

\author{Jiajie Fan\inst{1,2}\orcidlink{0009-0009-1825-5099} \and
Amal Trigui\inst{1}\orcidlink{0009-0001-7613-9872} \and
Thomas B\"ack\inst{2}\orcidlink{0000-0001-6768-1478} \and
Hao Wang\inst{2}\orcidlink{0000-0002-4933-5181}}

\authorrunning{J.Fan et al.}

\institute{BMW Group, Bremer Str.~6, 80788 Munich, Germany\\
\email{\{jiajie.fan, amal.trigui\}@bmw.de}\and
LIACS, Leiden University, Niels Bohrweg 1, 2333 CA Leiden, The Netherlands\\
\email{\{t.h.w.baeck,h.wang\}@liacs.leidenuniv.nl}
}
\maketitle

\begin{abstract}
A great interest has arisen in using Deep Generative Models (DGM) for generative design. When assessing the quality of the generated designs, human designers focus more on structural plausibility, \eg, no missing component, rather than visual artifacts, \eg, noises or blurriness. Meanwhile, commonly used metrics such as Fr\'echet Inception Distance (FID) may not evaluate accurately because they are sensitive to visual artifacts and tolerant to semantic errors. As such, FID might not be suitable to assess the performance of DGMs for a generative design task. In this work, we propose to encode the to-be-evaluated images with a Denoising Autoencoder (DAE) and measure the distribution distance in the resulting latent space. Hereby, we design a novel metric Fr\'echet Denoised Distance (FDD). We experimentally test our FDD, FID and other state-of-the-art metrics on multiple datasets, \eg, BIKED, Seeing3DChairs, FFHQ and ImageNet. Our FDD can effectively detect implausible structures and is more consistent with structural inspections by human experts. Our source code is publicly available at \url{https://github.com/jiajie96/FDD_pytorch}.
\keywords{Evaluation metric \and Generative design \and Structural plausibility}
\end{abstract}

\section{Introduction}
\label{sec:intro}
Following the swift development of Deep Generative Models (DGMs) in general image generation tasks~\cite{GAN14Ian,DCGAN,karras2019stylebased,ho2020ddpm}, a great interest has arisen in using DGMs to enable generative design~\cite{generativeDesign2021Regenwetter,CREATIVEGAN21Heyrani,fan2023adversarial, fan2023PoDM}, where DGMs are able to create innovative designs based on specific input requirements provided by users. In this particular domain, design data is responsible for representing the design object with structural and geometric patterns, which are required to be recognizable and plausible. In order to rank models during the development of generative models, recent works rely on a subjective evaluation~\cite{fan2023PoDM, Maiorca22FMD}, where human experts apply an established set of criteria to manually assess a significant quantity of generated data. This evaluation method yields reliable results, serving as ``ground truth'' for model ranking, but it is time-consuming and hard to reproduce~\cite{Maiorca22FMD}. Hence, for developing DGMs for design generation, it is necessary to have an automated metric, which is able to reliably quantify the goodness of the target DGM. 

Meanwhile, the evaluation of generated images is still an unsolved challenge among other general tasks in the DGM domain~\cite{Barratt2018ANO, naeem20metric, studyEval22}. DGM developers~\cite{Karras2017ProgressiveGO, Brock2018LargeSG, karras2019stylebased} are heavily relying on the Fr\'echet Inception Distance (FID)~\cite{heusel2018TTUR} metric, which extracts latent features from real and generated images with an Inception-V3~\cite{Szegedy2015inception} model pre-trained on ImageNet~\cite{DenDon09Imagenet} respectively and then quantifies their difference using Fr\'echet Distance as the final FID score. As the primary metric in the DGM field, FID is able to measure the fidelity and diversity and present them in a single value. However, a lot of studies~\cite{ProCons22FID, Stein23DinoFD, Kynk2023FID} disclose that FID does not always align with human evaluation and claim that this limitation is due to the reliance on the pre-trained Inception-V3 model. Hence, novel metrics are delivered by replacing the Inception-V3 model by other backbone networks, \eg, Clip,~\cite{Radford2021CLIP} VQ-VAE~\cite{van2017vqvae} and DINOv2~\cite{oquab2024dinov2}, \etc. According to the most recent work of Stein~\etal~\cite{Stein23DinoFD}, where they compared 17 metrics using encoders from 9 various networks, \DinoFD has the most reliable performance in terms of consistency with human judgment in their experiments.

\begin{figure}[t]
\fontsize{8}{10}\selectfont
\centering
\begin{subfigure}[b]{0.48\linewidth}
\centering
\includegraphics[width=\linewidth]{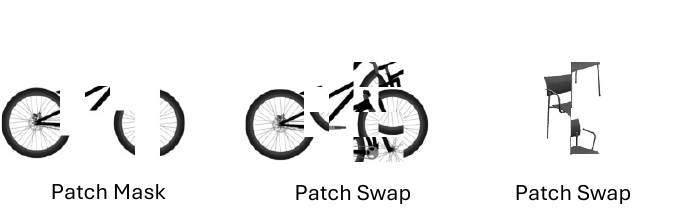}
\caption{Better scored by the SOTA metrics (\eg, FID, KID and \DinoFD)}
\end{subfigure}
\hfill
\unskip\ 
\vrule\ 
\hfill
\begin{subfigure}[b]{0.48\linewidth}
\centering
\includegraphics[width=\linewidth]{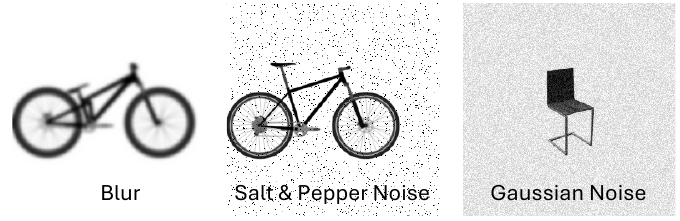}
\caption{Better scored by human designers and our FDD}
\end{subfigure}
\caption{\textbf{From which side (\textit{left} or \textit{right}) are the design images more plausible?} (a) Structural implausibility; (b) visual artifacts. Recent work~\cite{Baker2018DeepCN, Geirhos2018ImageNettrainedCA, Hermann2019TheOA} discovers that the SOTA metrics (FID, KID and \DinoFD) tend to penalize visual artifacts more than structural implausibility, which matters more to the human designers. In contrast, our FDD consists better with human designers and is able to focus on shapes.}
\label{fig:main_page} 
\end{figure}

On the other hand, recent works have pointed out that the Inception-V3 model and Inception-powered metrics perform poorly on shapes~\cite{Baker2018DeepCN, Geirhos2018ImageNettrainedCA, Hermann2019TheOA, fan2023PoDM}. Our work investigates this finding and observes that the state-of-the-art (SOTA) metrics generally suffer from this issue: they are sensitive to visual artifacts like noises, yet they have a high tolerance towards semantic failures, \eg, part missing in a bicycle, as illustrated in~\cref{fig:main_page}. Besides, human experts are able to recognize the same structural representation of the observed design image regardless of minor noise and they tend to penalize the evaluation based on the implausibility of the design more, rather than the presence of visual artifacts~\cite{Landau1988TheIO, Kucker2019ReproducibilityAA}. Motivated by this, our work aims to create a novel metric for generative design that is robust to visual corruption of the observed images and biased towards the design plausibility. 

Finally, we propose the Fr\'echet Denoised Distance (FDD) by replacing the Inception-V3 model within the FID framework with a Denoising Autoencoder (DAE)~\cite{DAE08Pascal} that has been also pre-trained on ImageNet dataset and capable of encoding images into latent features with an Inception-comparable dimension of $\mathbb{R}^{2048}$. The DAE is able to observe the same structural representation in the image regardless of the noisy disturbances, which can be utilized as a strong method to extract the structural feature from the noisy input. Our work compares our FDD with other SOTA metrics, \eg, FID, \DinoFD and Topology Distance (TD)~\cite{Horak21TD} (since their results show a similar bias to our intention), based on the following experiments: (1) sensitivity test over visual artifacts and structural failures; (2) consistency test with increasing disturbances; (3) consistency test with human judgment in model ranking. As a result, our FDD has the most stable performance among all the experiments. In order to explain the performance of FDD, we visualize the ``focus'' of our FDD metric compared to the FID using a GradCAM~\cite{Stein23DinoFD, GradCAM, Kynk2023FID} test, hereby showcasing that the DAE model has a better assessment regarding the requirements of human designers. In addition, our work explores the potential for further improvement of DAE-based metrics, where we build-upon concepts from existing works, \ie, KID~\cite{Binkowski18KID}, TD~\cite{Horak21TD} and training the network on structural images~\cite{Geirhos2018ImageNettrainedCA}, to design new DAE-based metrics, \ie, Kernel Denoised Distance (KDD), Topology Denoised Distance (TDD), and FDD ($\cdot$), respectively. We test these metrics on BIKED and the results show that these DAE-based metrics are highly correlated and FDD performs relatively best. 

\begin{figure}[t]
    \centering
    \includegraphics[width=\linewidth, trim=6mm 1mm 4mm 3mm, clip]{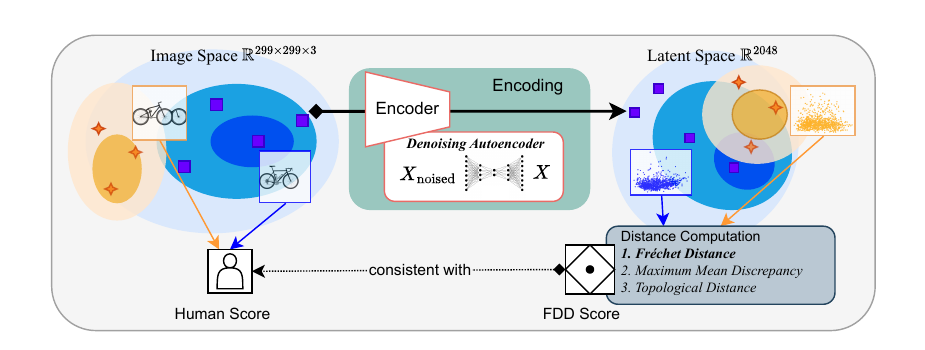}
    \caption{Plausibility evaluation using Fr\'echet Denoised Distance. \textit{Blue area and stars} visualizing the distribution and samples of real data in the image space and in the DAE-encoded latent space; \textit{Orange area and stars} illustrating the distribution and samples of generated data in the image space and in the DAE-encoded latent space.}
    \label{fig:FDD-Diagram}
\end{figure}

\section{Related Work}
Unlike the swift development in the field of Deep Generative Models (DGMs)~\cite{GAN14Ian, karras2019stylebased, dhariwal2021diffusion}, accurately ranking generative models remains an unresolved challenge~\cite{Barratt2018ANO, naeem20metric, studyEval22}. Humans are able to give the ground-truth evaluation in assessing a limited number of generated images, but quantifying the performance of a DGM requires an automated evaluation method~\cite{Stein23DinoFD}. Overcoming the flaws of previous metrics, \eg, SSIM~\cite{SSIM}, LPIPS~\cite{Zhang18Lpips} and IS~\cite{Salimans2016ImprovedTF}, currently most reported evaluation methods, \eg, Fr\'echet Inception Distance (FID)~\cite{heusel2018TTUR} and Kernel Inception Distance (KID)~\cite{Binkowski18KID}, have largely addressed the challenge of automated evaluation and are employed as the primary metric for model ranking in the field of DGM~\cite{Karras2017ProgressiveGO, Brock2018LargeSG, karras2019stylebased}. They leverage a two-step procedure: encode real and generated images into latent features in a lower-dimensional space with a representation extractor and then use a distance critic to quantify the difference between their features. Both FID and KID utilize the Inception-V3~\cite{Szegedy2015inception} model pre-trained on ImageNet, which has a 2048-dimensional latent space. Regarding the measurement of latent distance: FID fits the Inception features from real and generated images into a multivariate Gaussian before computing the Fr{\'e}chet Distance (also known as the Wasserstein-2 distance) between them; whereas KID~\cite{Binkowski18KID} uses the squared Maximum Mean Discrepancy (MMD)~\cite{gretton2012kernelMMD} with a polynomial kernel~\cite{studyEval22}.

Concerns about the over-reliance on the Inception-V3 model have been raised and researchers claim that an ImageNet~\cite{DenDon09Imagenet} classifier like the Inception-V3 model brings a significant bias to the evaluation with FID~\cite{naeem20metric, oquab2024dinov2}. Furthermore, FID is proved to be vulnerable to manipulation~\cite{Kynk2023FID}, especially when there exists a significant domain discrepancy between the data set of interest such as BIKED~\cite{regenwetter2021biked} and ImageNet~\cite{DenDon09Imagenet}. Consequently, the results measured by FID often show a poor correlation with human judgments. Similarly, KID~\cite{Binkowski18KID} encounters the same issue as it also leverages the pre-trained Inception-V3 model. Most recently, in order to find a perceptual representation space superior to the inception manifold, Stein~\etal~\cite{Stein23DinoFD} studied $17$ metrics with $9$ different encoders (\eg, CLIP~\cite{Radford2021CLIP}, SwAV~\cite{Mathilde20SwAV} and DINOv2~\cite{oquab2024dinov2}). Their finding concludes that \DinoFD~\cite{Stein23DinoFD} demonstrates the most reliable performance over various perspectives, \eg, fidelity, diversity, rarity, and memorization of generative models. Previous works~\cite{Baker2018DeepCN, Geirhos2018ImageNettrainedCA, Hermann2019TheOA} shed light on the role played by the image attributes, \eg, edges, shapes, textures, and colors in various computer vision tasks, \eg, classification and segmentation. 

Recent studies have introduced autoencoder-based metrics for evaluation purposes: for instance, Buzuti~\etal~\cite{buzuti2023frechet} leveraged the VQ-VAE~\cite{van2017vqvae} and showed that their unsupervised model-based metric Fréchet AutoEncoder Distance (FAED) outperforms FID in terms of consistency with increasing disturbance when evaluating on human and animal faces, \ie, CelebA HQ~\cite{liu2015deep}, FFHQ~\cite{karras2019stylebased}, and AFHQ~\cite{choi20AFHQ}. By cross-comparing their measured values among various types of disturbance, their FAED noticeably penalizes visual artifacts more severely than structural implausibility with comparable intensity. This may still lead to unfair comparisons of DGMs for design synthesis, where human experts prefer to use shape information for assessment~\cite{Landau1988TheIO, Kucker2019ReproducibilityAA}. Meanwhile, Horak~\etal~\cite{Horak21TD} proposed a more competitive shape-based evaluation metric by investigating the topological characteristics of potential flow shapes and proposed topological distance (TD) as a complementary metric for FID. Thus, we choose TD as for the later comparison.

\section{Method}
Considering the preference of human designers, the metric required by design generation should ``deprioritize'' the visual artifacts and instead focus on evaluating the underlying shape. To achieve this, we come to the idea of replacing the Inception-V3~\cite{Szegedy2015inception} model with the encoder of a trained Denoising Autoencoder (DAE)~\cite{DAE08Pascal}. Following this, our work introduces the Fr\'echet Denoised Distance (FDD).

\subsection{Preliminaries}

\label{sec: preliminaries}
\subsubsection{Fr\'echet Inception Distance (FID)}
The FID leverages the ImageNet-trained Inception-V3 model without its last fully connected layer. Hereby, it provides a lower-dimensional latent space. Real images $\vx$ and generated images $\vx^\prime$ are embedded into the Inception features $\vw\in\mathbb{R}^{2048}$ and $\vw^\prime\in\mathbb{R}^{2048}$, respectively, and then separately fitted into two multivariate Gaussian distributions, with $(\mu_{\vw}, \,\Sigma_{\vw})$ and $(\mu_{\vw^\prime}, \,\Sigma_{\vw^\prime})$ denoting the means and covariances thereof. The difference between the two latent manifolds will be quantified with Fr\'echet distance with:
\begin{equation}
 \label{eq:frechet_distance}
\text{FD}= {\|\mu_{\vw} - \mu_{\vw^\prime} \|}^2_2 + \Tr(\Sigma_{\vw}+ \Sigma_{\vw^\prime} - 2({\Sigma_{\vw}\Sigma_{\vw^\prime}})^{\frac{1}{2}}),
\end{equation}
where $\Tr(\cdot)$ computes the trace of a matrix.

\subsubsection{Denoising Autoencoder (DAE)}
The DAE~\cite{DAE08Pascal} is able to observe the same structural representation in the image regardless of the noisy disturbances, which demonstrates its robustness in assessing structural plausibility. The architecture of DAE is based on an expansion of the fundamental autoencoder model, consisting of two components: an encoder ($E_\theta \colon\vx\rightarrow\vw$) and a decoder ($D_\theta\colon\vw\rightarrow\vx$). In the training phase, source images $\vx\in\R^{w\times h\times c}$ are corrupted with Gaussian noises $\vx_\eta = \vx + \eta$, where $\eta \sim \mathcal{N}(\mathbf{0},\,\sigma^2 \cdot\mathbf{I})$ and $\sigma$ refers to the noise scale. The encoder ($E_\theta$) embeds the noised image $\vx_\eta$ into its lower-dimension latent representation $\vw = E_\theta(\vx_\eta)$, then the decoder restores the latent representation back into pixel-based image space $\hat{\vx} = D_\theta(\vw) = D_\theta\circ E_\theta(\vx_\eta)$. The network is trained using the following loss function:
\begin{equation}
\label{dae_loss}
   \min_{E_\theta, D_\theta}\Delta(\vx,\hat{\vx}) = \frac{1}{n}\sum_{i=1}^{n}(\vx_i-\hat{\vx}_i)^2 = \frac{1}{n}\sum_{i=1}^{n}(\vx_i-D_\theta\circ E_\theta(\vx_i + \eta))^2\;,
\end{equation}
where $n$ is the batch size.
While minimizing the reconstruction error in the training, Denoising Autoencoder (DAE) in turn maximizes the mutual information between the original input $\vx$ and its latent representation $\vw$~\cite{DAE08Pascal}. More specifically, DAE bypasses the noisy corruption between $\vx$ and $\hat{\vx}$. This allows the latent representation $\vw$ to contain meaningful information about the source $\vx$, even thought DAE only sees the corrupted input $\hat{\vx}$.


\subsection{Fr\'echet Denoised Distance (FDD)} 
We implement the encoder $E_\theta(\vx_\eta)$ of the Denoising Autoencoder (DAE) as the feature extractor. First, we design a DAE architecture (refer to~\cref{sec:configurations} for more information on this architecture) and train it on the ImageNet~\cite{DenDon09Imagenet} dataset with input shape of $299\times 299\times 3$.
Second, similarly to the procedure of FID, we embed a certain number $K$ of real images $\vx$ and generated images $\vx^\prime$ into the latent features $\vw\in\mathbb{R}^{2048}$ and $\vw^\prime\in\mathbb{R}^{2048}$, respectively. Note that the image is preprocessed into a shape of $299\times 299\times 3$ regardless of the original shape and color. Next, we follow the procedure of the Fr\'echet distance, introduced in~\cref{sec: preliminaries}, to quantify the difference between the two manifolds $\vw$ and $\vw^\prime$. Hereby, we design the Fr\'echet Denoised Distance (FDD), illustrated with an explanatory diagram in~\cref{fig:FDD-Diagram}. 

 For exploration purpose, we simulate the design processes of KID~\cite{Binkowski18KID} and TD~\cite{Horak21TD} and replace the distance measures with Maximum Mean Discrepancy (MMD) and Topology Distance (TD), hereby delivering more DAE-based metrics, \eg, Kernel Denoised Distance (KDD) and Topology Denoised Distance (TDD). We also notice the work of~\cite{Geirhos2018ImageNettrainedCA} that trains a ResNet-50~\cite{He16resnet} model on an alternative dataset of ImageNet, \ie, Stylized-ImageNet, and hereby successfully develops a shape-biased classifier. Inspired by this proposal, we additionally train a DAE model from scratch on the BIKED~\cite{regenwetter2021biked} dataset. The DAE model trained on BIKED images has an input shape of $256\times 256\times 1$ and a smaller latent space with dimension $D_{\vw} = 64$. Hereby, we design a FDD ($\cdot$) metric, which is based on the DAE trained on the same target dataset. The evaluation of FDD ($\cdot$) on BIKED images is demonstrated in~\cref{sec:Experiments}.

\section{Experiments}
\label{sec:Experiments}
To evaluate the design plausibility of generated images, a useful metric should satisfy the following conditions: (1) bias toward design structure, (2) consistency with increasing disturbances, and (3) alignment with human judgment. Hence, we leverage correspondingly three experiments: sensitivity test, consistency test with increasing disturbances, and model ranking, over the the SOTA metrics and our metrics (see~\cref{tab:evaluation_metrics} for more detailed information). Note that in our work, the TD metric refers to TD-Inception~\cite{Horak21TD} unless otherwise explained.

\subsection{Datasets}
\label{sec:datasets}

We select a variety of datasets covering different aspects. For a fair comparison with the FID, we train the DAE on the ImageNet~\cite{DenDon09Imagenet} dataset, ensuring that the learned feature manifold is similar to the one of the Inception-V3 model. We employ a subset of the ImageNet~\cite{DenDon09Imagenet} dataset of 50\,000 samples with dimension $299\times 299\times 3$, properly chosen to cover a wide range of 1\,000 classes. The dataset is divided into 45\,000 training samples and 5\,000 test samples. Our comparative analysis and tests also incorporate two design datasets, BIKED~\cite{regenwetter2021biked} and Seeing3DChairs~\cite{3Dchair}, to address the interests of human designers. Additionally, we incorporate the color-channeled FFHQ~\cite{karras2019stylebased} dataset and the test samples of ImageNet~\cite{DenDon09Imagenet} into our metric testing to confirm the metric's adaptability to general image generation tasks. Details of the implemented datasets for comparative analysis are provided in the supplementary material. 

\begin{table*}[t]
\caption{A list of candidate performance metrics for measuring design plausibility. Below the \textit{dashed line}, we also list other DAE-based metrics as a means of exploring further improvements. FDD ($\cdot$) utilizes a DAE trained on the target dataset with the DAE architecture modified according to the dataset, \eg, FDD (BIKED) utilizes a DAE trained on the BIKED dataset.}
\begin{center}\setlength{\tabcolsep}{2pt}
\renewcommand*{\arraystretch}{1.4}
\small
\resizebox{\textwidth}{!}{%
    \begin{tabular}{llcccl} 
    \hline
    \multirow{2}{*}{\textbf{Metric}} & \textbf{Backbone} & \textbf{Input} & \textbf{Feature} & \textbf{Training} & \textbf{Distance}\\
        & \textbf{Model} & \textbf{Dimension} & \textbf{Dimension} & \textbf{Dataset} & \textbf{Measures}\\\hline
    FID~\cite{heusel2018TTUR} &\multirow{2}{*}{Inception-V3~\cite{Szegedy2015inception}} & \multirow{2}{*}{$299\times 299\times 3$}  & \multirow{2}{*}{2048} & \multirow{2}{*}{ImageNet~\cite{DenDon09Imagenet}} & Fr\'echet Distance \\
    KID~\cite{Binkowski18KID} & & & &  & Maximum Mean Discrepancy \\\hline
    \DinoFD~\cite{Stein23DinoFD} &DINOv2~\cite{oquab2024dinov2}, ViT~\cite{dosovitskiy2020image}     & $224\times 224\times 3$ &     1024  & LVD-142M~\cite{Stein23DinoFD} & Fr\'echet Distance \\\hline
    TD-Inception~\cite{Horak21TD}&   Inception-V3~\cite{Szegedy2015inception} & $299\times 299\times 3$ & 2048& ImageNet~\cite{DenDon09Imagenet}  & \multirow{2}{*}{Topology Distance} \\
    TD-ResNet~\cite{Horak21TD} &ResNet18~\cite{He16resnet}& $224\times 224\times 3$ & 512 &  Fashion-MNIST~\cite{Xiao2017FashionMNISTAN}& \\
    \hline
    \textbf{FDD} & Denoising Autoencoder~\cite{DAE08Pascal} & $299\times 299\times 3$& 2048& ImageNet~\cite{DenDon09Imagenet} &  Fr\'echet Distance \\
    \hdashline
    KDD& \multirow{3}{*}{Denoising Autoencoder~\cite{DAE08Pascal}}  & \multirow{2}{*}{$299\times 299\times 3$}&  \multirow{2}{*}{2048}& \multirow{2}{*}{ImageNet~\cite{DenDon09Imagenet}} & Maximum Mean Discrepancy \\
    TDD& & & & & Topology Distance  \\
    FDD ($\cdot$)& & $256\times 256\times 1$ & 64 &  Target Dataset &  Fr\'echet Distance \\\bottomrule
    \end{tabular}
}
\end{center}
\label{tab:evaluation_metrics}
\end{table*}
\subsection{Experimental Settings}
\label{sec:configurations}
For the reproducibility of our work, this section documents all the essential details regarding the development of our FDD metric and the experimental setups. To justify the setting choices, we aim to align our DAE model's architecture with that of the Inception-V3 model, particularly in terms of input shape and latent dimension. The model architecture and training settings describe the DAE trained on ImageNet, whereas the configurations of the DAE trained on BIKED are correspondingly adjusted as shown in~\cref{tab:evaluation_metrics}.

\subsubsection{Model Architecture}
Our approach employs a DAE comprising $5$ convolutional layers across both the encoder and the decoder. Here, the feature dimensions for the convolutional layers in the encoder are arranged in the following sequence $[32, 64, 128, 256, 512]$. For the decoder, these dimensions are applied in reverse order. Each layer employs a $3\times 3$ kernel shape, a stride of $2$, padding of $1$, and the \texttt{Rectified Linear Unit (ReLU)} as the activation function, aligning with the Inception-V3 model. The last activation layer of the decoder uses a \texttt{Tanh} function to adjust the outputs to a pixel range of $[-1,1]$. In alignment with the configuration parameters of the Inception-V3 model, the encoder's input shape is specified as $299\times 299\times 3$, and the latent vector dimension is established at $2048$. See the supplementary material for visualization of the model architecture.

\subsubsection{Training Settings}
The training process uses a subset of 45\,000 images from ImageNet~\cite{DenDon09Imagenet}, which are rescaled to the range $[-1,1]$. For the DAE training set-up, input images are corrupted with Gaussian noise $\mathcal{N}(\mathbf{0},\,\sigma^2 \cdot\mathbf{I})$ with $\sigma = 0.1$ before being fed into the encoder. We utilize the \texttt{Adam} Optimizer with a learning rate of $1\mathrm{e}{-3}$ to train the DAE with a batch size of $128$ and epochs of $1\,000$. The reconstruction loss is assessed by calculating the mean squared error (MSE) between the original and output images. Model performance is continuously assessed during training, and the best-performing model is chosen from the saved checkpoints for further experiments. We also implement an early stop function, where the training stops if the reconstruction loss does not reduce within $20$ epochs. In the supplementary material, we provide qualitative results of reconstructing images from various datasets using the DAE trained on ImageNet.

\subsubsection{Disturbance Procedures} For conducting the sensitivity and consistency tests that exam metrics' performance in dealing with various disturbances, we design the perturbation methods, \ie, salt \& pepper noise, Gaussian noise, patch mask, patch swap and a mixed disturbance of Gaussian noise and patch swap, and their respective intensity levels based on previous studies~\cite{heusel2018TTUR, Horak21TD}. Details of the disturbances are provided in the supplementary material.

\begin{figure}[t]
    \centering
    \centering
    \includegraphics[width=\linewidth]{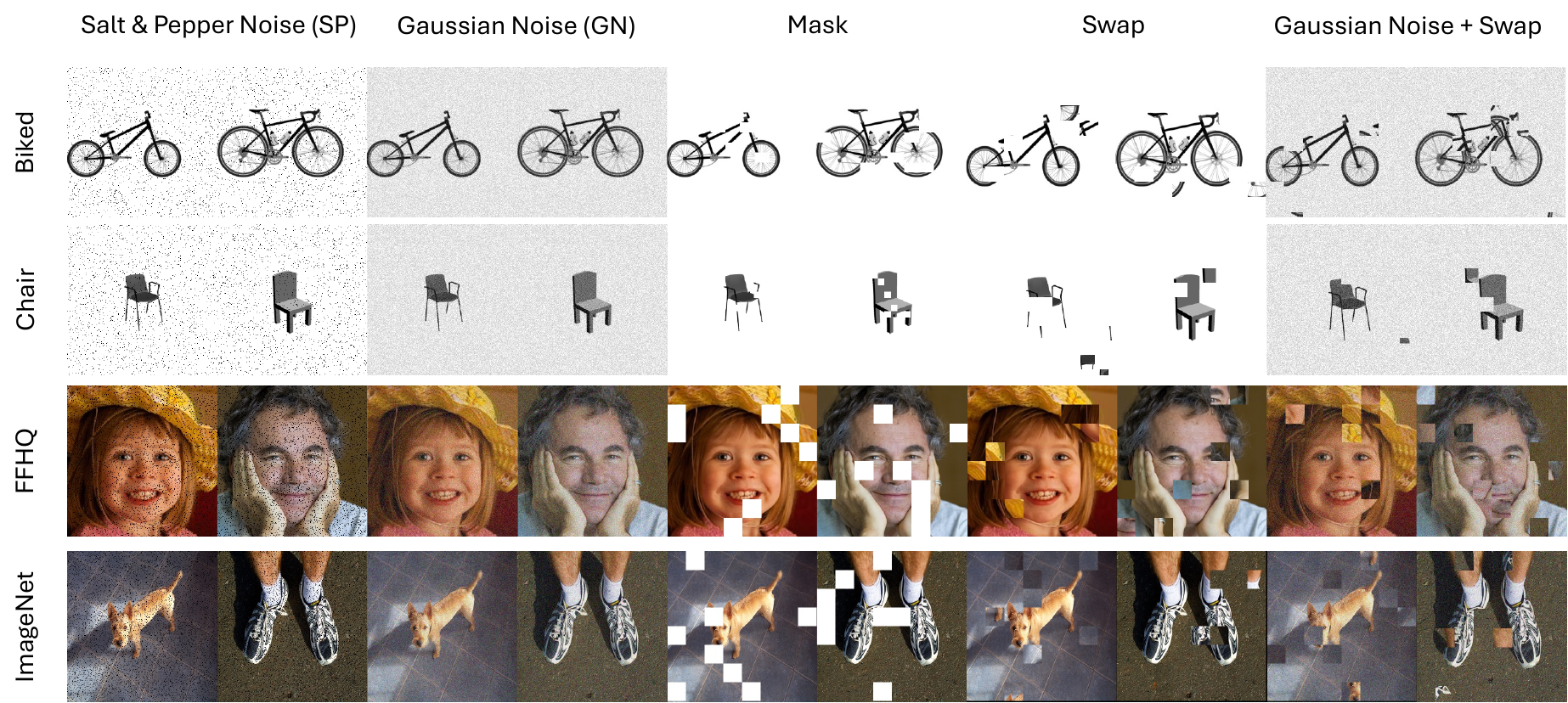}
    \caption{Examples of manipulated images for sensitivity test. We choose the intensity of the disturbances so that images with structural errors (\ie, mask and swap) are notably less plausible than ones with visual artifacts (\ie, salt \& pepper noise and Gaussian noise).}
    \label{fig:disturbed_images_samples}
\end{figure}

\begin{figure}[t]
\centering
\begin{subfigure}[b]{0.49\linewidth}
        \centering
    \includegraphics[width=\linewidth]{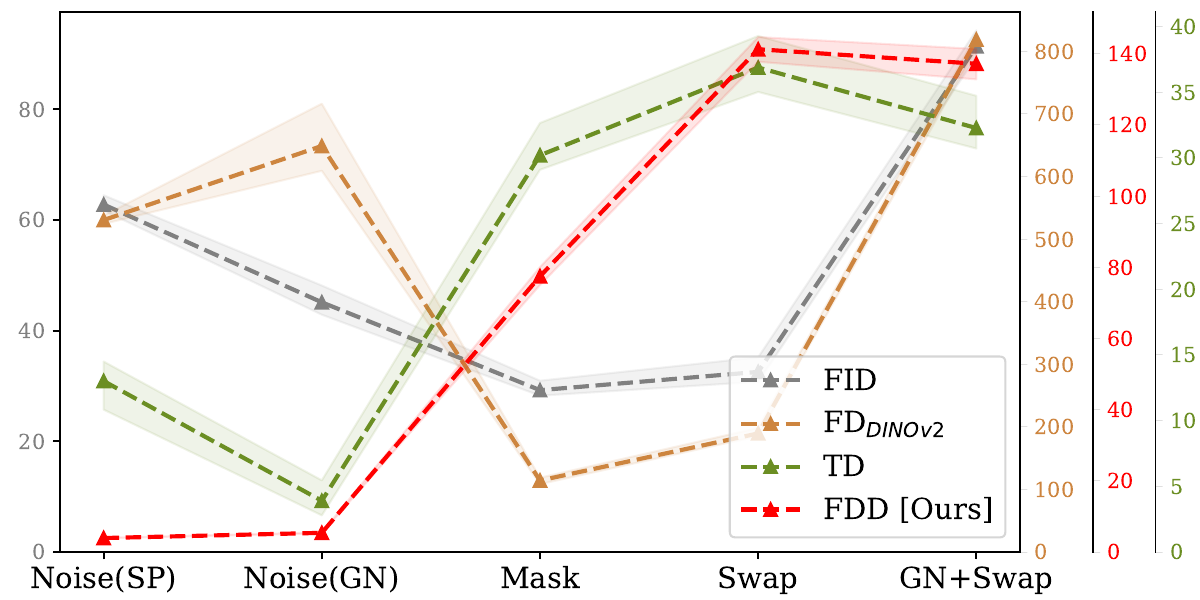}
    \caption{On BIKED~\cite{regenwetter2021biked}}
    \label{fig:sensitivity_test_Biked}
\end{subfigure}
\hfill
\begin{subfigure}[b]{0.49\linewidth}
        \centering
    \includegraphics[width=\linewidth]{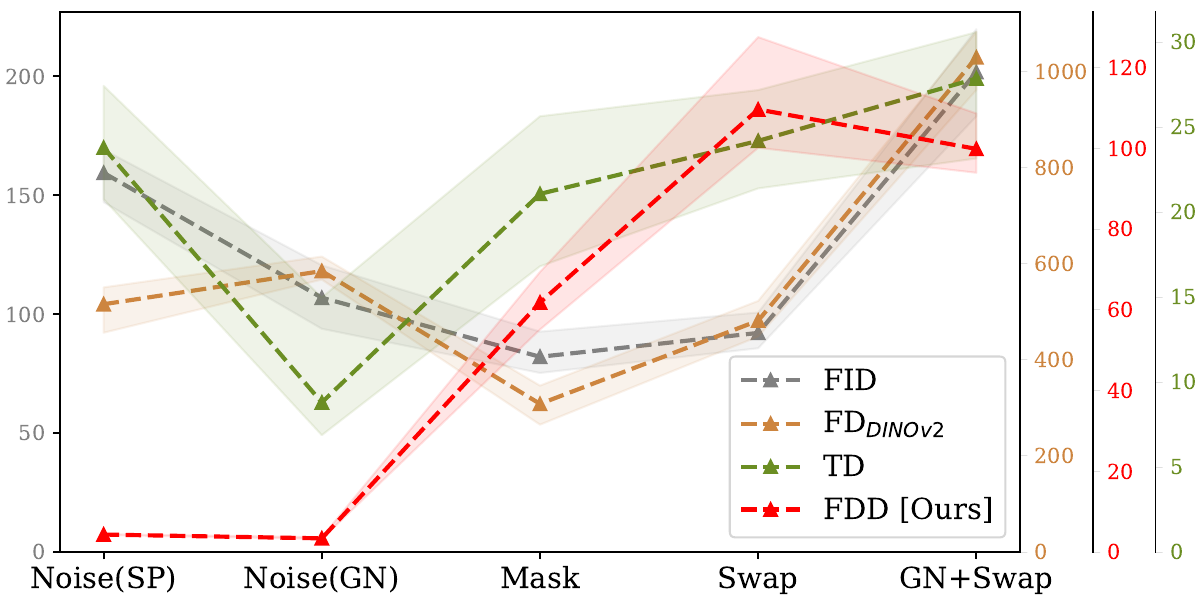}
    \caption{On Seeing3DChairs~\cite{3Dchair}}
    \label{fig:sensitivity_test_chair}
\end{subfigure}\\

\begin{subfigure}[b]{0.49\linewidth}
        \centering
    \includegraphics[width=\linewidth]{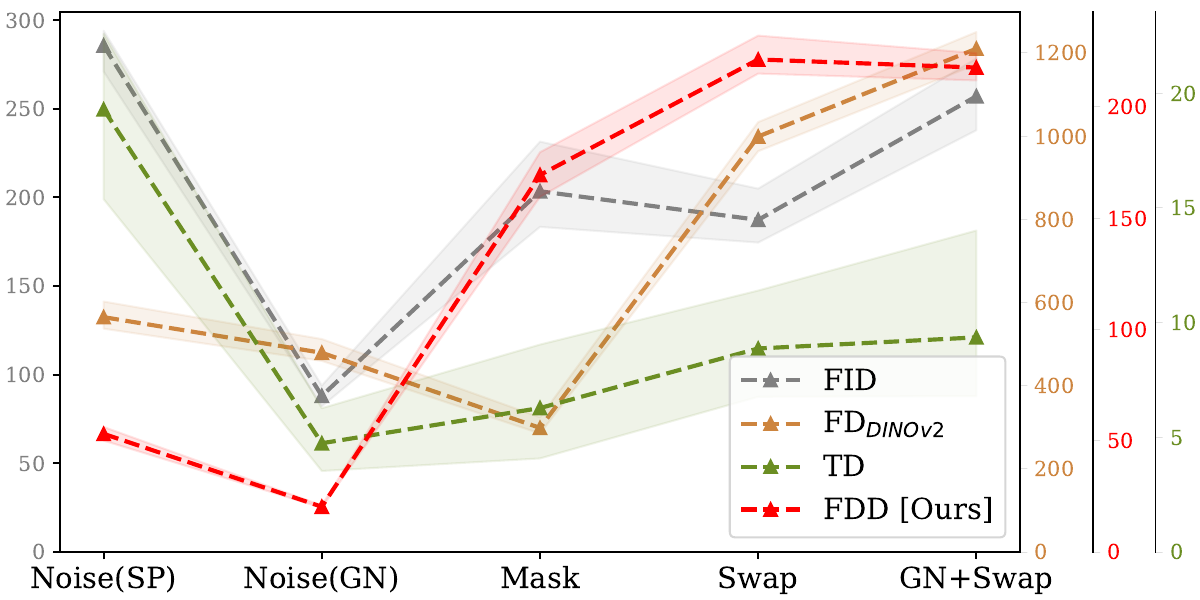}
    \caption{On FFHQ~\cite{karras2019stylebased}}
    \label{fig:sensitivity_test_ffhq}
\end{subfigure}
\hfill
\begin{subfigure}[b]{0.49\linewidth}
        \centering
    \includegraphics[width=\linewidth]{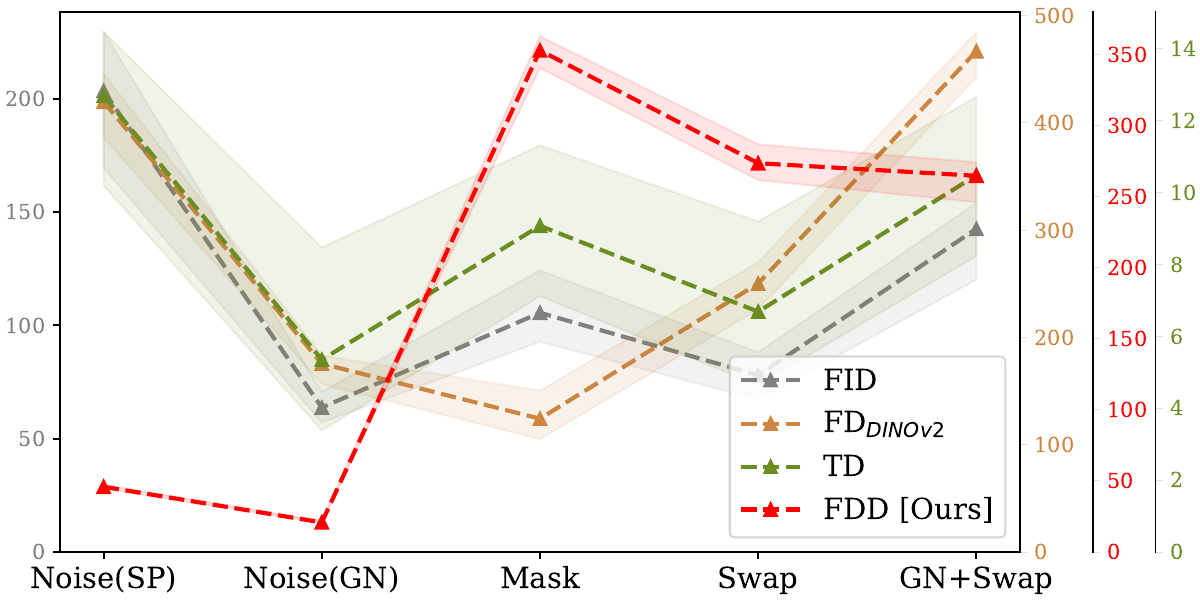}
    \caption{On ImageNet~\cite{DenDon09Imagenet}}
    \label{fig:sensitivity_test_imageNet}
\end{subfigure}
\caption{Sensitivity Comparison. The y-axis represents the score value measured by each metric, where a lower value indicates a higher similarity to source data, \ie, better quality. A reliable plausibility metric should penalize more on the basis of structural errors (\eg, mask and swap) than visual artifacts (\eg, noise). For each metric, the \textit{dashed line} shows the mean across the groups and the \textit{shaded region} depicts the measured values from the groups.}
\label{fig:sensitivity_comparison}
\end{figure}

\begin{figure}[t]
\centering
\begin{subfigure}[t]{0.62\linewidth}
    \centering
\includegraphics[width=\linewidth]{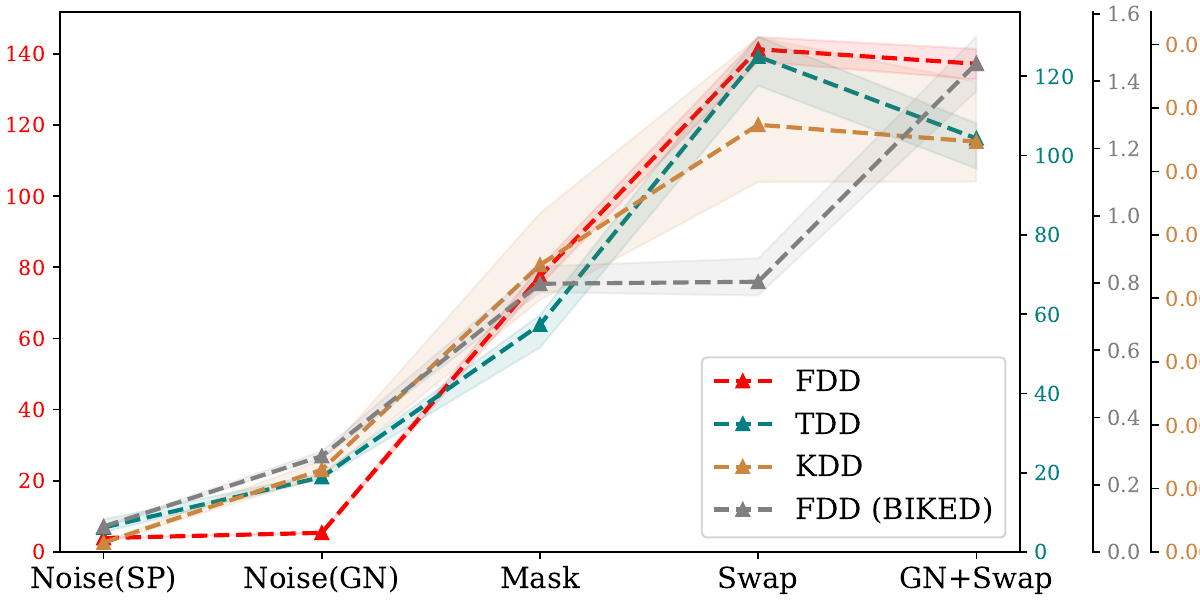}
\caption{Sensitivity test on BIKED~\cite{regenwetter2021biked}}
\label{fig:mutations_FDD}
\end{subfigure}
\hfill
\begin{subfigure}[t]{0.36\linewidth}
\centering
\includegraphics[width=\linewidth, trim=28mm 0mm 40mm 8mm,clip]{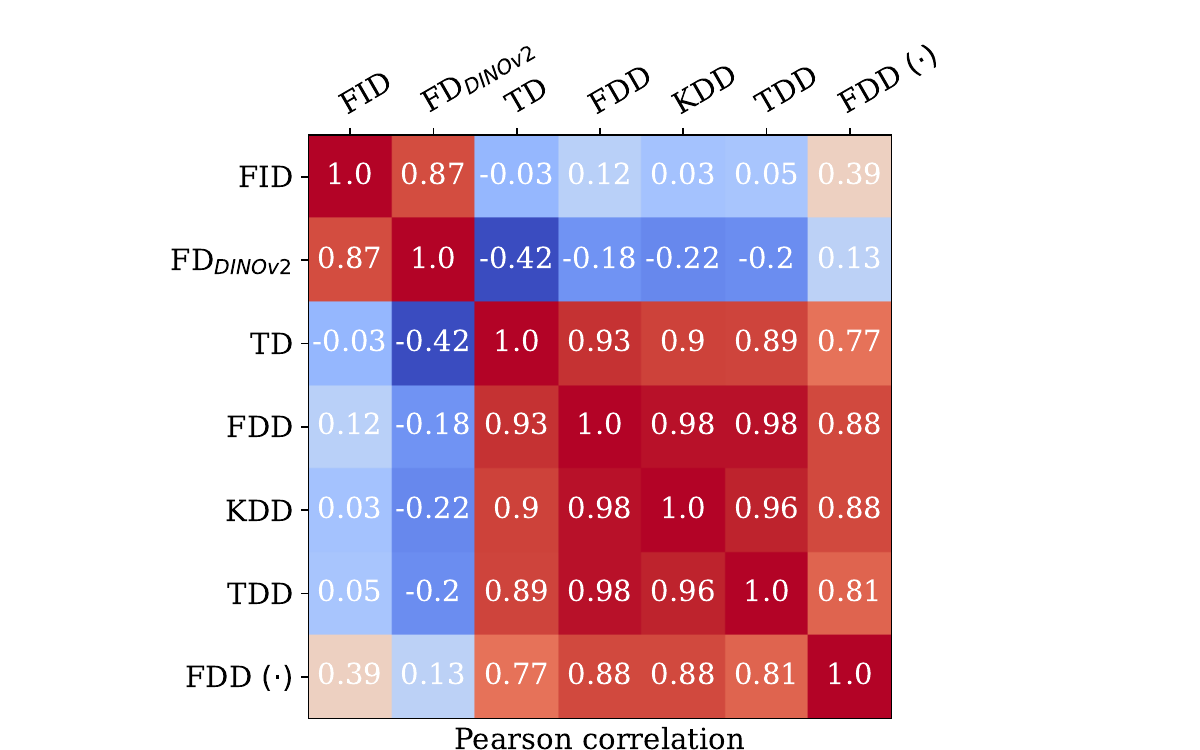}
\caption{Correlation test}
\label{fig:Pearsonr_coeff_Biked}
\end{subfigure}
\caption{Experiments with FDD and other DAE-based metrics. In (a), within all DAE-based metrics, FDD shows the best performance; (b) Pearson correlation of metrics over all distances measured during the sensitivity test with BIKED.}
\label{fig:sensitivity_comparison_FDD_mutations}
\end{figure}

\subsection{Sensitivity Test}
\label{sec:Sensitivity-comparison}

Despite the presence of noise, a human designer can still recognize the underlying structure in a design. However, designs with missing parts or structural errors are less usable. Thus, we design the sensitivity test with the anticipation that an appropriate metric for the design generation evaluation task should progressively demonstrate deteriorating scores from visual artifacts to structural deficiencies. Additionally, to prove the importance of structural integrity in the evaluation process, we expect that the score for a mixed disturbance of Gaussian noise and patch swap will be comparable to that of solely patch swap disturbance, thus remaining independent from the added visual artifacts. The aim of the sensitivity test is to cross-compare the metric performance in dealing with various disturbances and to see if the metric aligns with human designers. 

This test involves four datasets: BIKED~\cite{regenwetter2021biked}, ImageNet~\cite{DenDon09Imagenet}, FFHQ~\cite{karras2019stylebased} and Seeing3DChairs~\cite{3Dchair}. For each dataset, we shuffle and split the samples into $n = 10$ groups, number of samples in each group varies from the dataset: $K=300$ (for BIKED) and $K=100$ (for Seeing3DChairs, FFHQ and ImageNet). Considering the effect of sample number to the metric performance, we provide results of sensitivity test with 1k images for each dataset in the supplementary material. We introduce five types of disturbances into source images and create five corrupted counterparts, \ie, pepper noise, Gaussian noise, patch mask, patch swap and a mix of Gaussian noise and patch swap. The introduced disturbances adhere to a rule where visual artifacts, such as pepper noise and Gaussian noise, are intentionally kept at levels that do not significantly impact the recognition of the design. On the other hand, structural failures, such as patch masking and patch swapping, lead to designs that are implausible and consequently receive worse human evaluation scores compared to visual artifacts. We choose one level from each disturbance described in~\cref{sec:configurations}: $\alpha=0.01\,(\text{pepper noise, Gaussian noise}), \text{and} \, \alpha=0.25\,(\text{patch mask, \text{patch swap}})$. Next, we measure the distance between each one of these corrupted image sets and the original image set, using FID, \DinoFD, TD, and our FDD. Since they are measures of distance quantifying the dissimilarity between observed images and source images, a smaller value indicates greater similarity to the source data.

We plot several examples of disturbed images in~\cref{fig:disturbed_images_samples} and record the measured results in~\cref{fig:sensitivity_comparison}. As expected, FID~\cite{heusel2018TTUR} and~\DinoFD~\cite{Stein23DinoFD} show a great bias towards visual artifacts, with notably higher distance assigned to pepper and Gaussian noised images compared to those with patch mask and patch swap. In contrast to FID and \DinoFD, TD and our FDD provide a distinct evaluation perspective by detecting structural faults and imposing penalties accordingly. One unanticipated result was that TD exhibits a poor performance with regard to pepper noise as illustrated in~\cref{fig:sensitivity_test_chair} and \cref{fig:sensitivity_test_ffhq}. Furthermore, as the sample size decreases within each group from $300$ (BIKED~\cite{regenwetter2021biked}) to $100$ (for Seeing3DChairs~\cite{3Dchair} and FFHQ~\cite{karras2019stylebased}), TD shows a significant increase in standard deviation across $10$-times implementations. Interestingly, our FDD shows a better stability among various noises and gives significantly worse scores to images with structural failures.

Furthermore, we explore other possible metrics based on DAE by incorporating concepts from existing works such as KID~\cite{Binkowski18KID}, TD~\cite{Horak21TD} and training the network on structural images~\cite{Geirhos2018ImageNettrainedCA}.
This adaption yields new evaluation metrics, \ie, Kernel Denoised Distance (KDD), Topology Denoised Distance (TDD), and FDD (BIKED), respectively. Later on, we subject these metrics to the sensitivity test and present the results in~\cref{fig:mutations_FDD}. Our analysis reveals that FDD exhibits the most consistent performance across various criteria: the most stable result across different groups and excellence in distinguishing between visual and structural disturbances.

Finally, we calculate the Pearson correlation coefficients pair-wisely among all candidate metrics, by taking the measured values from~\cref{fig:sensitivity_test_Biked}, and record the outcome in the table~\cref{fig:Pearsonr_coeff_Biked}. Notably, the result reveals two categories among the metrics: FID and \DinoFD are grouped together, while TD and our designed metrics demonstrate a stronger correlation with each other. This is a promising finding since TD's main perspective is the topology and geometric behavior of the latent space, hereby we argue that the latent space of our DAE maintains the topological properties of the image space well and can be captured by Fr\'echet Distance. Note that the KDD, TDD and FDD (BIKED) are experimental explorations. They are highly related to our FDD in the correlation test and our FDD outperforms them in the sensitivity test.

\begin{figure}[t]
    \centering
    \includegraphics[width=0.98\linewidth]{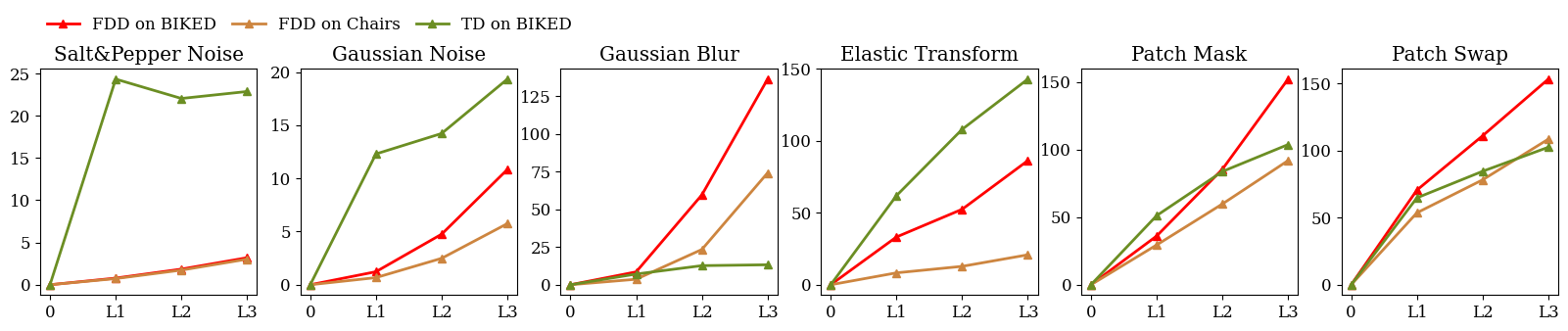}
    \caption{Metric comparison with increasing disturbances. The y-axis label represents the measured distance of each metric. The TD, as the most competitive metric to our FDD, performs however unstably with increasing disturbances.}
    \label{fig:increasing_disturbances}
\end{figure}

\subsection{Consistency with Increasing Disturbances}
\label{sec:increasing_disturbances}
In this section, we test the consistency of the FDD metric in response to escalating levels of disturbances outlined in~\cref{sec:configurations}. As a fundamental requirement, a performance metric should be able to accurately detect and respond to worsened image quality, including visual fidelity and structural plausibility. We start by adding various disturbances to a subset comprising $K=1\,000$ images sourced from the BIKED~\cite{regenwetter2021biked} and Seeing3DChairs~\cite{3Dchair} datasets, respectively. Afterwards, we report the scores in~\cref{fig:increasing_disturbances} and demonstrate the consistent performance of the proposed FDD metric. While FID has been noted to exhibit inconsistency in detecting the disturbance level induced by salt and pepper as documented in Heusel~\etal~\cite{heusel2018TTUR}, our FDD successfully measures the levels of various deformations, spanning from visual to structural distortions.

\begin{figure}[t]
    \centering
    \includegraphics[width=\linewidth, trim=12mm 0mm 2mm 5mm,clip]{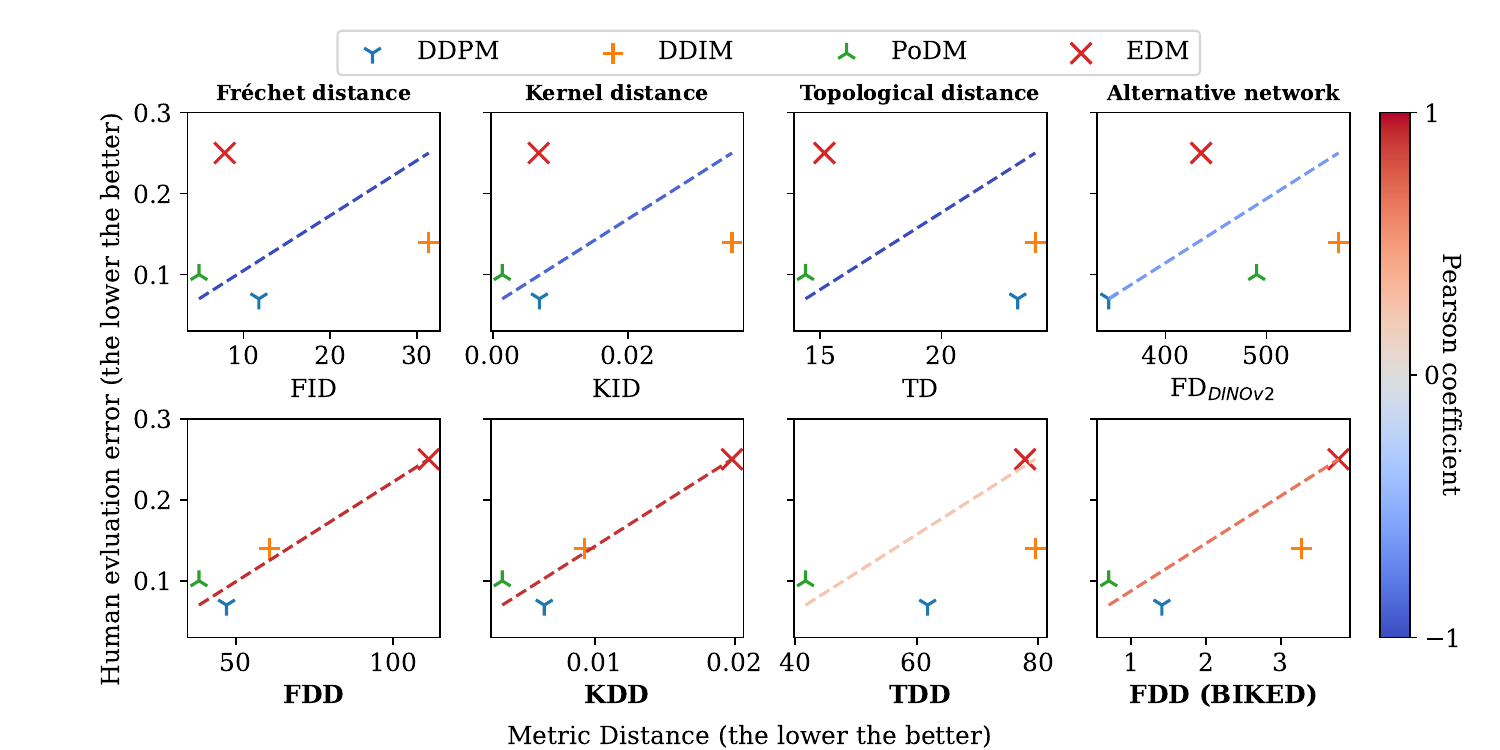}
    \caption{Metrics comparison in the task of model ranking. We color-code the \textit{diagonal lines} after the measured Pearson correlation coefficient between metric results and human judgments, \textit{dark red} refers to a strong positive correlation between metric distances and human judgments. }
    \label{fig:ranking_Biked}
\end{figure}

\begin{figure}[t]
    \centering
    \begin{subfigure}[b]{\linewidth}
            \centering
        \includegraphics[width=\linewidth]{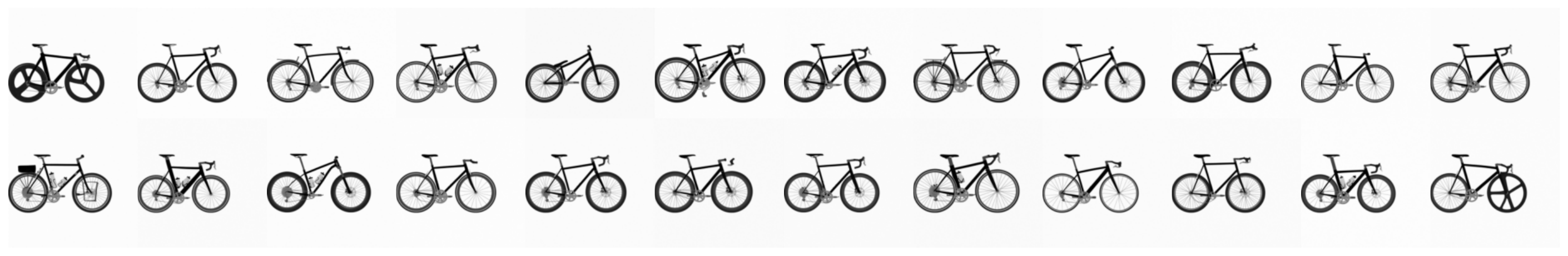}
        \caption{DDPM~\cite{ho2020ddpm} (FID: $11.77$, \DinoFD: $342.82$, FDD: $48.08$)}
    \end{subfigure}
    \begin{subfigure}[b]{\linewidth}
            \centering
        \includegraphics[width=\linewidth]{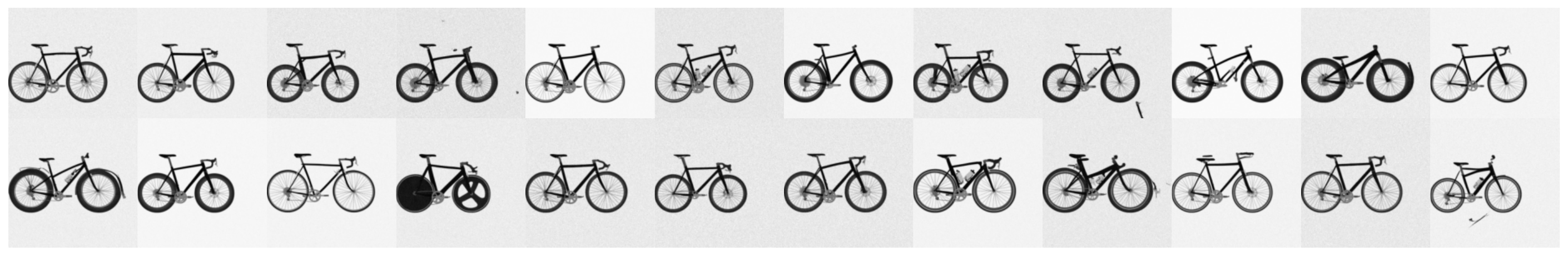}
        \caption{DDIM~\cite{song2022ddim} (FID: $31.35$, \DinoFD: $571.21$, FDD: $60.66$)}
    \end{subfigure}
    \begin{subfigure}[b]{\linewidth}
        \centering
        \includegraphics[width=\linewidth]{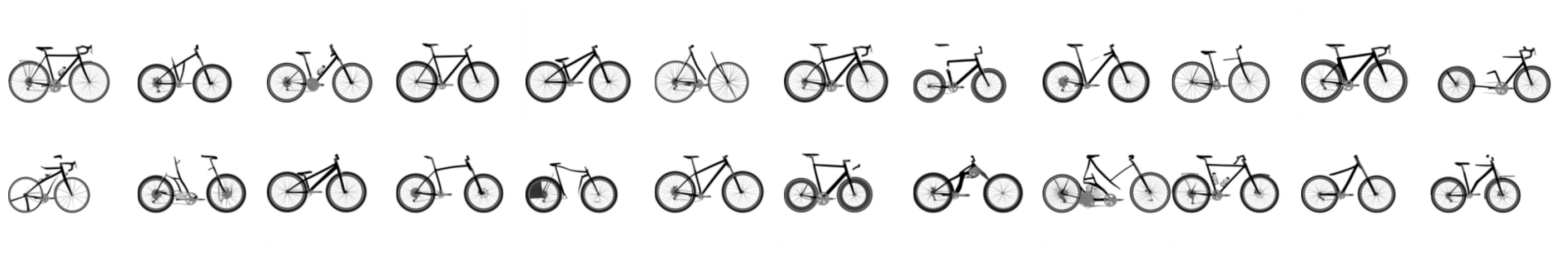}
        \caption{EDM~\cite{karras2022edm} (FID: $7.84$, \DinoFD: $435.25$, FDD: $111.25$)}
    \end{subfigure}

    \caption{Qualitative evaluation of generated bicycle designs. DDPM and DDIM yield structurally more plausible results than EDM, but FID and \DinoFD fail to agree with human judgments, whereas our FDD ranks the models with the perspective of structural plausibility and penalizes the visual artifacts as well.}
    \label{fig:generated_Biked}
\end{figure}

\subsection{Model Ranking}
\label{sec:Ranking}
In the model ranking, we employ five deep generative models, \eg, DDPM~\cite{ho2020ddpm}, DDIM~\cite{song2022ddim}, EDM~\cite{karras2022edm} and PoDM~\cite{fan2023PoDM}, with the consideration that the models executed in model ranking should exhibit significant differences in visual quality and structural plausibility. These models are then trained on BIKED images with a resolution of $256\times 256$. We generate 5k images from each model and manually evaluate them into plausible designs and implausible designs. We denote the ratio of implausible bicycle designs as human evaluation error, the lower the better, which serves as the ``ground truth'' in this model ranking experiment.

Meanwhile, we apply the candidate metrics, including FID, KID, \DinoFD, TD, FDD, and other DAE-based metrics, to evaluate each generative model with their generated samples, with 1k images in each group. Subsequently, the distances measured and human error rates are visualized in~\cref{fig:ranking_Biked}. Note that the proximity of the plotted points (measured distances, human evaluation error) to the diagonal line signifies the consistency of the metric with human evaluation. Particularly, the expected behavior is seen in the FDD, KDD, and FDD (BIKED) measurements, which are highly associated and yield the same consistent ranking. This observation aligns with the notion proposed by~\cite{Stein23DinoFD}, whereby provided a good encoder is chosen, all these metrics provide sensible ways of quantifying distances between probability distributions. 

On the other hand, the absence of a significant link between the SOTA metrics and human evaluation suggests a deficiency of these most reported metrics in evaluating structural design images. In~\cref{fig:generated_Biked}, we plot the generated bicycles for qualitative evaluation of our FDD metric. EDM achieves the best FID of 7.84, but the generated bicycles contain a large portion of implausible designs; DDPM (FID $11.77$) and DDIM (FID $31.35$) are unfairly penalized, even though the results are significantly more plausible than those from EDM. Here, our FDD is able to rank the models more accurately.

\begin{figure}[t]
\fontsize{8}{10}\selectfont
\centering
\begin{subfigure}[b]{0.49\linewidth}
\centering
\includegraphics[width=\linewidth, trim=50mm 20mm 45mm 30mm,clip]{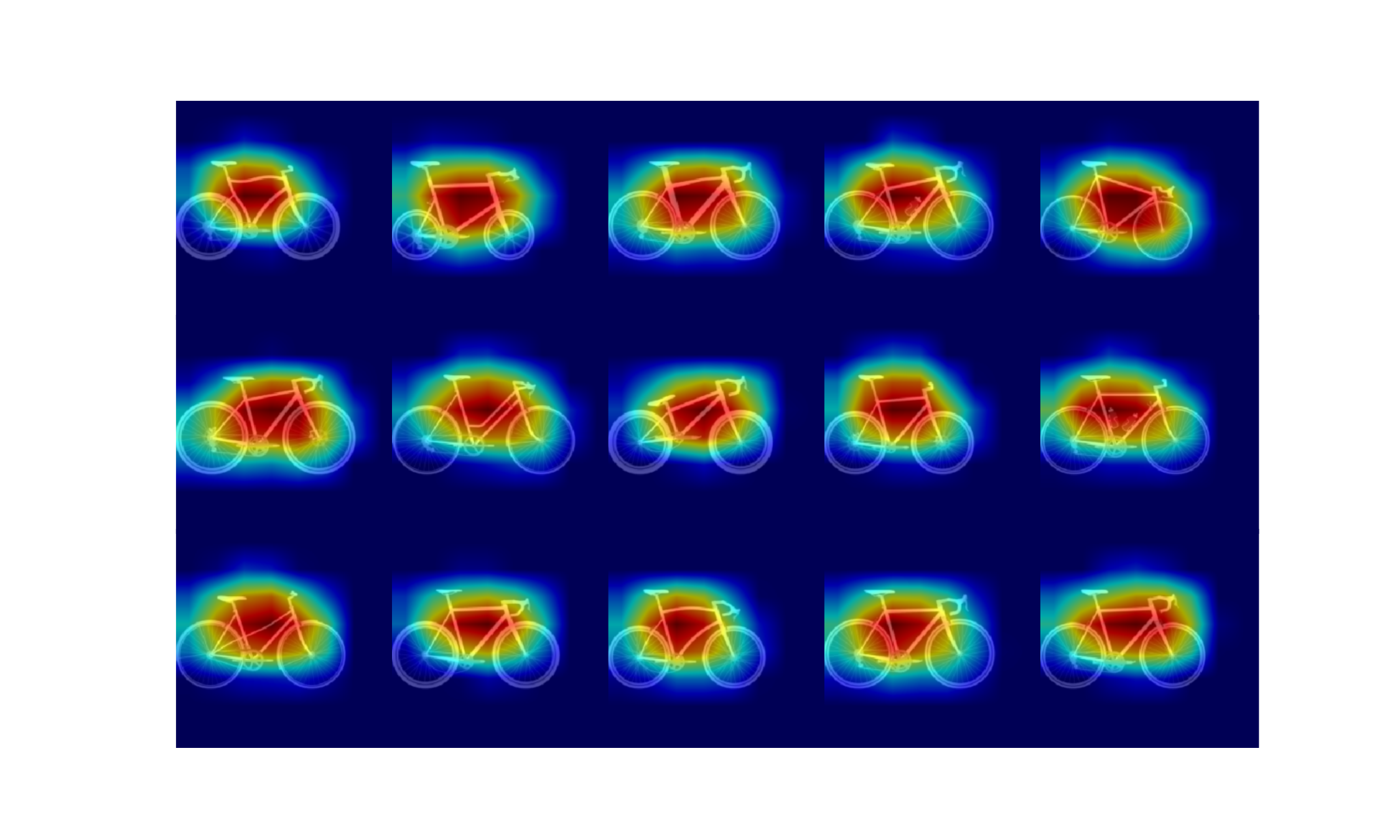}%
\caption{FID (Inception-V3)}
\label{fig:latent_inception}
\end{subfigure}
\hfill
\begin{subfigure}[b]{0.49\linewidth}
\centering
\includegraphics[width=\linewidth, trim=50mm 20mm 45mm 30mm,clip]{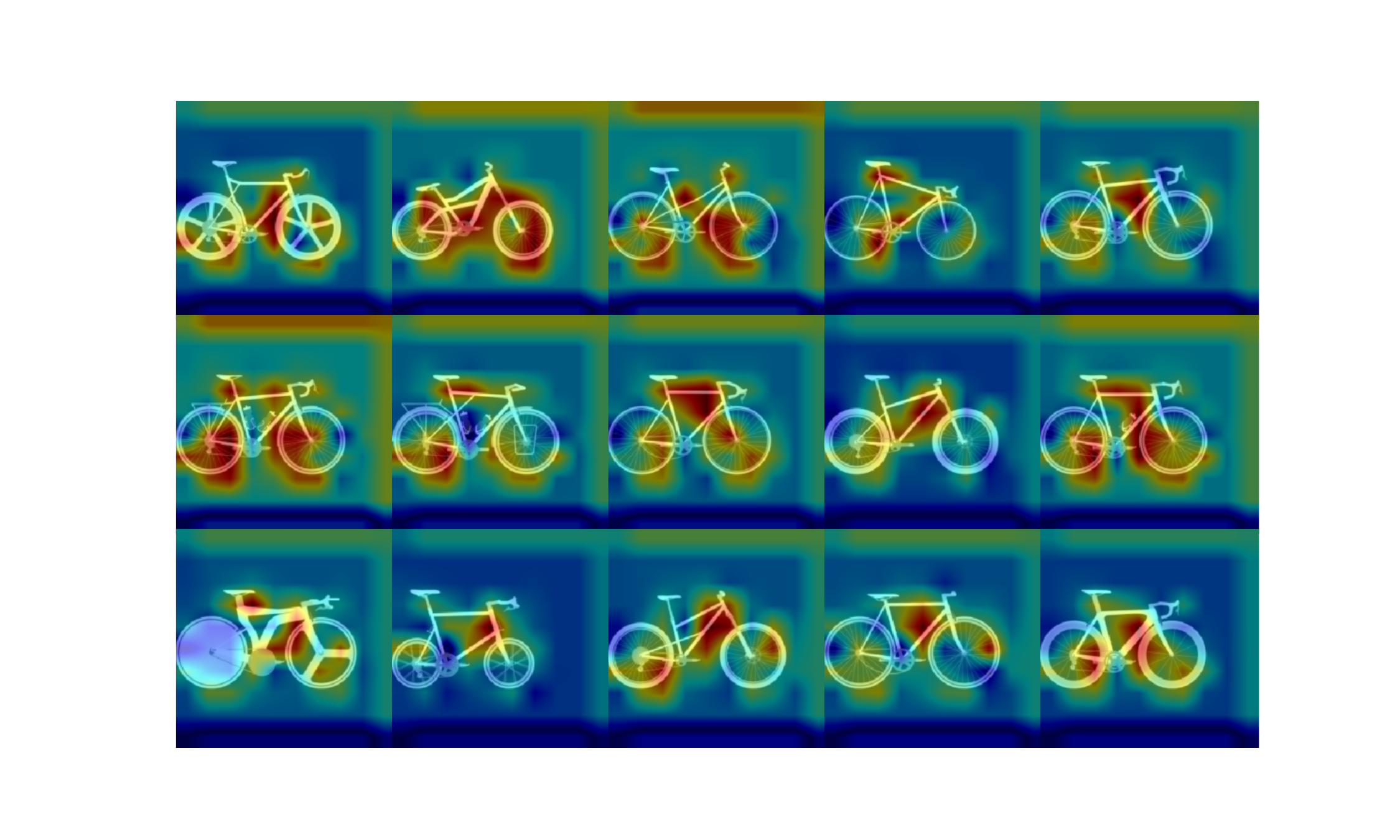}%
\caption{Our FDD (DAE trained on ImageNet)}
\label{fig:latent_DAE}
\end{subfigure}
\caption{\textbf{Where does the metric look at?} Heatmaps illustrating the perception of the Fr\'echet distance with Grad-CAM. The focus of an encoder can be demonstrated by both \textit{bright red} and \textit{deep blue}. The offset of the focusing area can be caused by upsampling the attention map to the image shape.}
\label{fig:latent_distribution_maps} 
\end{figure}

\subsection{Grad-CAM Visualization}
\label{sec:Grad-CAM-test}

The Grad-CAM~\cite{GradCAM, liu2020towards} is designed to visualize the focus on the input image as perceived by the classifier/segmentation model up to the last fully connected layer. In our work, we use the Grad-CAM visualization to compare the observation fields of the FID and FDD metrics. We first transfer the test images into inception space and latent space via the Inception-V3 model and the DAE model, respectively, which have the same dimension of $\mathbb{R}^{2048}$. We compute the mean $\mu_{\vw}$ and the covariance $\Sigma_{\vw}$ of the extracted features. Then, we obtain the attention maps of FID and FDD by back-propagating the value of ${\mu_{\vw}}^2+\Sigma_{\vw}$, to the last convolutional layer of the Inception-V3 model (\ie, \textit{Mixed\,7c.branch\,pool}) and the one of DAE (\ie, \textit{encoder\,8}), respectively. The Grad-CAM generates a heatmap of reduced dimensions (\eg, $10 \times 10$ for DAE) which is then upsampled to match the dimensions of the original image for intuitive visual comparison. The heatmaps (seen in~\cref{fig:latent_distribution_maps}) visualize the area observed by the corresponding metric in the BIKED~\cite{regenwetter2021biked} images, \ie, where the metric ``looks at'' when it calculates the distance. 

The focus of the Inception-V3 model is simply the area around the center of the main object, often mismatching the object's shape and borders. As explained in previous works~\cite{Stein23DinoFD, Kynk2023FID}, this phenomenon is caused by the model's classification training across 1\,000 classes. Consequently, it prioritizes detecting the object's presence rather than its structure. On the other hand, even when also trained on ImageNet, the DAE generates an intensive attention map with positive and negative gradients surrounding the bicycle's structure, which efficiently assesses the complex details of the bicycle's shape. In supplementary material, we provide also Grad-CAM analysis on general images, where Inception-V3 model tends to drop structural information while DAE captures it.

\section{Conclusion}
In this work, we approached the field of evaluating generated design images and proposed a structure-biased metric Fr\'echet Denoised Distance (FDD) by replacing the Inception-V3 model in the FID metric with a Denoising Autoencoder, unsupervised-trained on the same dataset (\ie, ImageNet) and with the same 2048-dimensional latent space. Through a series of experiments, including sensitivity test for various types of disturbance, consistency test with increasing disturbances, and alignment test with human judgment in model ranking, we found FDD to fulfill the quality requirements for serving as a metric and outperform other SOTA metrics, \eg, FID, \DinoFD and TD, on design images such as BIKED and Seeing3DChairs, as well as real-world images such as human faces from FFHQ and general images from ImageNet. We explained the effectiveness of FDD with a Grad-CAM visualization, where the DAE is able to ``focus'' on the design structure of the observed shape.

\subsubsection{Limitation and Future Work} FDD has a low priority towards visual artifacts, which may become problematic in contexts where the visual quality is critical. Yet, our experiments demonstrated that when there is no major structural failure, it is still capable of penalizing visual artifacts. In addition, we believe that this novel insight may also be useful in guiding DGMs to generate more reliable, plausible designs, which is of great potential to be further investigated in future work. Finally, while our research on FDD mainly focused on image space, considering the recent study of 3D Diffusion modeling (adding noise to 3D data) and the use of an autoencoders to process 3D data, we believe that FDD can also assess 3D-DGMs.

\section*{Acknowledgements}
We gratefully acknowledge Laure Vuaille for her valuable insights and the support provided by BMW Group.

%
%
\bibliographystyle{splncs04}

\begin{thebibliography}{10}
\providecommand{\url}[1]{\texttt{#1}}
\providecommand{\urlprefix}{URL }
\providecommand{\doi}[1]{https://doi.org/#1}

\bibitem{3Dchair}
Aubry, M., Maturana, D., Efros, A.A., Russell, B.C., Sivic, J.: Seeing 3d chairs: Exemplar part-based 2d-3d alignment using a large dataset of cad models. In: 2014 IEEE CVPR. pp. 3762--3769 (2014). \doi{10.1109/CVPR.2014.487}

\bibitem{Baker2018DeepCN}
Baker, N., Lu, H., Erlikhman, G., Kellman, P.J.: Deep convolutional networks do not classify based on global object shape. PLoS Computational Biology  \textbf{14} (2018), \url{https://api.semanticscholar.org/CorpusID:54476941}

\bibitem{Barratt2018ANO}
Barratt, S.T., Sharma, R.: A note on the inception score. ArXiv  \textbf{abs/1801.01973} (2018), \url{https://api.semanticscholar.org/CorpusID:38384342}

\bibitem{studyEval22}
Betzalel, E., Penso, C., Navon, A., Fetaya, E.: A study on the evaluation of generative models. CoRR  \textbf{abs/2206.10935} (2022). \doi{10.48550/ARXIV.2206.10935}, \url{https://doi.org/10.48550/arXiv.2206.10935}

\bibitem{Binkowski18KID}
Binkowski, M., Sutherland, D.J., Arbel, M., Gretton, A.: Demystifying {MMD} gans. In: 6th International Conference on Learning Representations, {ICLR} 2018, Vancouver, BC, Canada, April 30 - May 3, 2018, Conference Track Proceedings. OpenReview.net (2018), \url{https://openreview.net/forum?id=r1lUOzWCW}

\bibitem{ProCons22FID}
Borji, A.: Pros and cons of gan evaluation measures: New developments. Computer Vision and Image Understanding  \textbf{215},  103329 (2022). \doi{https://doi.org/10.1016/j.cviu.2021.103329}, \url{https://www.sciencedirect.com/science/article/pii/S1077314221001685}

\bibitem{Brock2018LargeSG}
Brock, A., Donahue, J., Simonyan, K.: Large scale gan training for high fidelity natural image synthesis. ArXiv  \textbf{abs/1809.11096} (2018), \url{https://api.semanticscholar.org/CorpusID:52889459}

\bibitem{buzuti2023frechet}
Buzuti, L.F., Thomaz, C.E.: Fr{\'e}chet autoencoder distance: A new approach for evaluation of generative adversarial networks. Computer Vision and Image Understanding  \textbf{235},  103768 (2023)

\bibitem{Mathilde20SwAV}
Caron, M., Misra, I., Mairal, J., Goyal, P., Bojanowski, P., Joulin, A.: Unsupervised learning of visual features by contrasting cluster assignments. In: Proceedings of the 34th International Conference on Neural Information Processing Systems. NIPS'20, Curran Associates Inc., Red Hook, NY, USA (2020)

\bibitem{choi20AFHQ}
Choi, Y., Uh, Y., Yoo, J., Ha, J.W.: Stargan v2: Diverse image synthesis for multiple domains. In: 2020 IEEE/CVF Conference on Computer Vision and Pattern Recognition (CVPR). pp. 8185--8194 (2020). \doi{10.1109/CVPR42600.2020.00821}

\bibitem{DenDon09Imagenet}
Deng, J., Dong, W., Socher, R., Li, L.J., Li, K., Fei-Fei, L.: Imagenet: A large-scale hierarchical image database. In: Computer Vision and Pattern Recognition, 2009. CVPR 2009. IEEE Conference on. pp. 248--255. IEEE (2009), \url{https://ieeexplore.ieee.org/abstract/document/5206848/}

\bibitem{dhariwal2021diffusion}
Dhariwal, P., Nichol, A.Q.: Diffusion models beat {GAN}s on image synthesis. In: Beygelzimer, A., Dauphin, Y., Liang, P., Vaughan, J.W. (eds.) Advances in Neural Information Processing Systems (2021), \url{https://openreview.net/forum?id=AAWuCvzaVt}

\bibitem{dosovitskiy2020image}
Dosovitskiy, A., Beyer, L., Kolesnikov, A., Weissenborn, D., Zhai, X., Unterthiner, T., Dehghani, M., Minderer, M., Heigold, G., Gelly, S., et~al.: An image is worth 16x16 words: Transformers for image recognition at scale. arXiv preprint arXiv:2010.11929  (2020)

\bibitem{fan2023PoDM}
Fan, J., Vuaille, L., Bäck, T., Wang, H.: On the noise scheduling for generating plausible designs with diffusion models (2023)

\bibitem{fan2023adversarial}
Fan, J., Vuaille, L., Wang, H., Bäck, T.: Adversarial latent autoencoder with self-attention for structural image synthesis. arXiv preprint arXiv:2307.10166  (2023)

\bibitem{Geirhos2018ImageNettrainedCA}
Geirhos, R., Rubisch, P., Michaelis, C., Bethge, M., Wichmann, F., Brendel, W.: Imagenet-trained cnns are biased towards texture; increasing shape bias improves accuracy and robustness. ArXiv  \textbf{abs/1811.12231} (2018), \url{https://api.semanticscholar.org/CorpusID:54101493}

\bibitem{GAN14Ian}
Goodfellow, I., Pouget-Abadie, J., Mirza, M., Xu, B., Warde-Farley, D., Ozair, S., Courville, A., Bengio, Y.: Generative adversarial nets. In: Advances in neural information processing systems. pp. 2672--2680 (2014), \url{http://papers.nips.cc/paper/5423-generative-adversarial-nets.pdf}

\bibitem{gretton2012kernelMMD}
Gretton, A., Borgwardt, K.M., Rasch, M.J., Sch{\"o}lkopf, B., Smola, A.: A kernel two-sample test. The Journal of Machine Learning Research  \textbf{13}(1),  723--773 (2012)

\bibitem{He16resnet}
He, K., Zhang, X., Ren, S., Sun, J.: Deep residual learning for image recognition. In: 2016 IEEE Conference on Computer Vision and Pattern Recognition (CVPR). pp. 770--778 (2016). \doi{10.1109/CVPR.2016.90}

\bibitem{Hermann2019TheOA}
Hermann, K.L., Chen, T., Kornblith, S.: The origins and prevalence of texture bias in convolutional neural networks. arXiv: Computer Vision and Pattern Recognition  (2019), \url{https://api.semanticscholar.org/CorpusID:220266152}

\bibitem{heusel2018TTUR}
Heusel, M., Ramsauer, H., Unterthiner, T., Nessler, B., Hochreiter, S.: Gans trained by a two time-scale update rule converge to a local nash equilibrium. Advances in Neural Information Processing Systems 30 (NIPS 2017)  (2018)

\bibitem{ho2020ddpm}
Ho, J., Jain, A., Abbeel, P.: Denoising diffusion probabilistic models. arXiv preprint arxiv:2006.11239  (2020)

\bibitem{Horak21TD}
Horak, D., Yu, S., Khorshidi, G.S.: Topology distance: {A} topology-based approach for evaluating generative adversarial networks. In: Thirty-Fifth {AAAI} Conference on Artificial Intelligence, {AAAI} 2021, Thirty-Third Conference on Innovative Applications of Artificial Intelligence, {IAAI} 2021, The Eleventh Symposium on Educational Advances in Artificial Intelligence, {EAAI} 2021, Virtual Event, February 2-9, 2021. pp. 7721--7728. {AAAI} Press (2021). \doi{10.1609/AAAI.V35I9.16943}, \url{https://doi.org/10.1609/aaai.v35i9.16943}

\bibitem{Karras2017ProgressiveGO}
Karras, T., Aila, T., Laine, S., Lehtinen, J.: Progressive growing of gans for improved quality, stability, and variation. ArXiv  \textbf{abs/1710.10196} (2017), \url{https://api.semanticscholar.org/CorpusID:3568073}

\bibitem{karras2022edm}
Karras, T., Aittala, M., Aila, T., Laine, S.: Elucidating the design space of diffusion-based generative models. Advances in Neural Information Processing Systems  \textbf{35},  26565--26577 (2022)

\bibitem{karras2019stylebased}
Karras, T., Laine, S., Aila, T.: A style-based generator architecture for generative adversarial networks. In: Proceedings of the IEEE/CVF conference on computer vision and pattern recognition. pp. 4401--4410 (2019)

\bibitem{Kucker2019ReproducibilityAA}
Kucker, S.C., Samuelson, L.K., Perry, L.K., Yoshida, H., Colunga, E., Lorenz, M.G., Smith, L.B.: Reproducibility and a unifying explanation: Lessons from the shape bias. Infant behavior \& development  \textbf{54},  156--165 (2019), \url{https://api.semanticscholar.org/CorpusID:53045726}

\bibitem{Kynk2023FID}
Kynk{\"a}{\"a}nniemi, T., Karras, T., Aittala, M., Aila, T., Lehtinen, J.: The role of imagenet classes in fr\'echet inception distance. In: The Eleventh International Conference on Learning Representations (2023), \url{https://openreview.net/forum?id=4oXTQ6m_ws8}

\bibitem{Landau1988TheIO}
Landau, B., Smith, L.B., Jones, S.S.: The importance of shape in early lexical learning. Cognitive Development  \textbf{3},  299--321 (1988), \url{https://api.semanticscholar.org/CorpusID:205117480}

\bibitem{liu2020towards}
Liu, W., Li, R., Zheng, M., Karanam, S., Wu, Z., Bhanu, B., Radke, R.J., Camps, O.: Towards visually explaining variational autoencoders. In: Proceedings of the IEEE/CVF Conference on Computer Vision and Pattern Recognition. pp. 8642--8651 (2020)

\bibitem{liu2015deep}
Liu, Z., Luo, P., Wang, X., Tang, X.: Deep learning face attributes in the wild. In: Proceedings of the IEEE international conference on computer vision. pp. 3730--3738 (2015)

\bibitem{Maiorca22FMD}
Maiorca, A., Yoon, Y., Dutoit, T.: Evaluating the quality of a synthesized motion with the fr\'{e}chet motion distance. In: ACM SIGGRAPH 2022 Posters. SIGGRAPH '22, Association for Computing Machinery, New York, NY, USA (2022). \doi{10.1145/3532719.3543228}, \url{https://doi.org/10.1145/3532719.3543228}

\bibitem{naeem20metric}
Naeem, M.F., Oh, S.J., Uh, Y., Choi, Y., Yoo, J.: Reliable fidelity and diversity metrics for generative models. In: III, H.D., Singh, A. (eds.) Proceedings of the 37th International Conference on Machine Learning. Proceedings of Machine Learning Research, vol.~119, pp. 7176--7185. PMLR (13--18 Jul 2020), \url{https://proceedings.mlr.press/v119/naeem20a.html}

\bibitem{CREATIVEGAN21Heyrani}
Nobari, A.H., Rashad, M.F., Ahmed, F.: Creativegan: Editing generative adversarial networks for creative design synthesis. CoRR  \textbf{abs/2103.06242} (2021), \url{https://arxiv.org/abs/2103.06242}

\bibitem{oquab2024dinov2}
Oquab, M., Darcet, T., Moutakanni, T., Vo, H.V., Szafraniec, M., Khalidov, V., Fernandez, P., HAZIZA, D., Massa, F., El-Nouby, A., Assran, M., Ballas, N., Galuba, W., Howes, R., Huang, P.Y., Li, S.W., Misra, I., Rabbat, M., Sharma, V., Synnaeve, G., Xu, H., Jegou, H., Mairal, J., Labatut, P., Joulin, A., Bojanowski, P.: {DINO}v2: Learning robust visual features without supervision. Transactions on Machine Learning Research  (2024), \url{https://openreview.net/forum?id=a68SUt6zFt}

\bibitem{Radford2021CLIP}
Radford, A., Kim, J.W., Hallacy, C., Ramesh, A., Goh, G., Agarwal, S., Sastry, G., Askell, A., Mishkin, P., Clark, J., Krueger, G., Sutskever, I.: Learning transferable visual models from natural language supervision. In: International Conference on Machine Learning (2021), \url{https://api.semanticscholar.org/CorpusID:231591445}

\bibitem{DCGAN}
Radford, A., Metz, L., Chintala, S.: Unsupervised representation learning with deep convolutional generative adversarial networks. In: Bengio, Y., LeCun, Y. (eds.) 4th International Conference on Learning Representations, {ICLR} 2016, San Juan, Puerto Rico, May 2-4, 2016, Conference Track Proceedings (2016), \url{http://arxiv.org/abs/1511.06434}

\bibitem{regenwetter2021biked}
Regenwetter, L., Curry, B., Ahmed, F.: {BIKED: A Dataset for Computational Bicycle Design With Machine Learning Benchmarks}. Journal of Mechanical Design  \textbf{144}(3) (10 2021). \doi{10.1115/1.4052585}, \url{https://doi.org/10.1115/1.4052585}, 031706

\bibitem{generativeDesign2021Regenwetter}
Regenwetter, L., Nobari, A.H., Ahmed, F.: Deep generative models in engineering design: A review. Journal of Mechanical Design  \textbf{144}(7),  071704 (2022)

\bibitem{Salimans2016ImprovedTF}
Salimans, T., Goodfellow, I.J., Zaremba, W., Cheung, V., Radford, A., Chen, X.: Improved techniques for training gans. ArXiv  \textbf{abs/1606.03498} (2016), \url{https://api.semanticscholar.org/CorpusID:1687220}

\bibitem{GradCAM}
Selvaraju, R.R., Cogswell, M., Das, A., Vedantam, R., Parikh, D., Batra, D.: Grad-cam: Visual explanations from deep networks via gradient-based localization. In: 2017 IEEE International Conference on Computer Vision (ICCV). pp. 618--626 (2017). \doi{10.1109/ICCV.2017.74}

\bibitem{song2022ddim}
Song, J., Meng, C., Ermon, S.: Denoising diffusion implicit models. arXiv:2010.02502  (October 2020), \url{https://arxiv.org/abs/2010.02502}

\bibitem{Stein23DinoFD}
Stein, G., Cresswell, J.C., Hosseinzadeh, R., Sui, Y., Ross, B.L., Villecroze, V., Liu, Z., Caterini, A.L., Taylor, J.E.T., Loaiza{-}Ganem, G.: Exposing flaws of generative model evaluation metrics and their unfair treatment of diffusion models. CoRR  \textbf{abs/2306.04675} (2023). \doi{10.48550/ARXIV.2306.04675}, \url{https://doi.org/10.48550/arXiv.2306.04675}

\bibitem{Szegedy2015inception}
Szegedy, C., Vanhoucke, V., Ioffe, S., Shlens, J., Wojna, Z.: Rethinking the inception architecture for computer vision. 2016 IEEE Conference on Computer Vision and Pattern Recognition (CVPR) pp. 2818--2826 (2015), \url{https://api.semanticscholar.org/CorpusID:206593880}

\bibitem{van2017vqvae}
Van Den~Oord, A., Vinyals, O., et~al.: Neural discrete representation learning. Advances in neural information processing systems  \textbf{30} (2017)

\bibitem{DAE08Pascal}
Vincent, P., Larochelle, H., Bengio, Y., Manzagol, P.A.: Extracting and composing robust features with denoising autoencoders. In: International Conference on Machine Learning (2008), \url{https://api.semanticscholar.org/CorpusID:207168299}

\bibitem{Xiao2017FashionMNISTAN}
Xiao, H., Rasul, K., Vollgraf, R.: Fashion-mnist: a novel image dataset for benchmarking machine learning algorithms. ArXiv  \textbf{abs/1708.07747} (2017), \url{https://api.semanticscholar.org/CorpusID:702279}

\bibitem{Zhang18Lpips}
Zhang, R., Isola, P., Efros, A.A., Shechtman, E., Wang, O.: The unreasonable effectiveness of deep features as a perceptual metric. In: 2018 {IEEE} Conference on Computer Vision and Pattern Recognition, {CVPR} 2018, Salt Lake City, UT, USA, June 18-22, 2018. pp. 586--595. Computer Vision Foundation / {IEEE} Computer Society (2018). \doi{10.1109/CVPR.2018.00068}, \url{http://openaccess.thecvf.com/content\_cvpr\_2018/html/Zhang\_The\_Unreasonable\_Effectiveness\_CVPR\_2018\_paper.html}

\bibitem{SSIM}
Zhou, W.: Image quality assessment: from error measurement to structural similarity. IEEE transactions on image processing  \textbf{13},  600--613 (2004)

\end{thebibliography}

\end{document}